# Achieving "Synergy" in Cognitive Behavior of Humanoids via Deep Learning of Dynamic Visuo-Motor-Attentional Coordination

Jungsik Hwang[1], Minju Jung[1], Naveen Madapana[2], Jinhyung Kim[1], Minkyu Choi[1] and Jun Tani[1*]

[1]Korea Advanced Institute of Science and Technology, Daejeon, South Korea
`{jungsik.hwang, minju5436, kkjh0723, minkyu.choi8904, tani1216jp}@gmail.com`
[2]Indian Institute of Technology, Guwahati, India
`naveen.madapana@gmail.com`

**Abstract.** The current study examines how adequate coordination among different cognitive processes including visual recognition, attention switching, action preparation and generation can be developed via learning of robots by introducing a novel model, the Visuo-Motor Deep Dynamic Neural Network (VMDNN). The proposed model is built on coupling of a dynamic vision network, a motor generation network, and a higher level network allocated on top of these two. The simulation experiments using the iCub simulator were conducted for cognitive tasks including visual object manipulation responding to human gestures. The results showed that "synergetic" coordination can be developed via iterative learning through the whole network when spatio-temporal hierarchy and temporal one can be self-organized in the visual pathway and in the motor pathway, respectively, such that the higher level can manipulate them with abstraction.

**Keywords:** Artificial Intelligence, Learning, Robotics

## 1   Introduction

It is desired that humanoid robots can learn to generate complex goal-directed behaviors with using dynamic visual image patterns as like human does. In many situations, generation of such behaviors involves with multiple cognitive processes. For example, let us consider a visually-guided goal-directed action of robots in the following. A robot initially pays attention to its human partner who specifies a target object through a demonstrative gesture, and then the robot shifts its visual attention to the specified object. The robot recognizes the object's shape, location and orientation and reaches its hand to the target object with pre-shaping based on the visual information attained. At the same time, the robot avoids collision to obstacles by switching attention to them as necessary. Finally, the robot grasps and lifts up the object. Mastering this sort of skilled goal-directed actions requires adequate coordination among a set of cogni-tive processes including visual recognition, attention switching, action preparation and generation. It is essential to link these basic cognitive processes with synergy by developing spatio-temporal coordination among them. The current paper proposes that such synergy can be naturally developed in terms of dynamical structure in a particular configuration of "deep" neural network models via consolidative learning of visuo-motor experience of robots.

Recently, deep learning has attracted significant attention in its potential capability for learning to extract latent structure out of massive amount of exemplar patterns. It has been shown that some of deep learning schemes outperformed conventional pattern matching methods in accuracy which use hand-coded features. For example, Convolutional Neural Network (CNN) outperformed the conventional vision recognition schemes in categorization capability for static visual objects [1]. In the field of speech recognition, it was reported that a version of Recurrent Neural Network (RNN) outperformed conventional scheme based on Hidden Markov Model (HMM) [2].

It is naturally expected that introduction of deep learning schemes such as CNN and RNN could contribute significantly to improvement of learning capability in humanoid robots if they are adequately applied to the problems. On this account, the current paper proposes a novel model, named as the Visuo-Motor Deep Dynamic Neural Network (VMDNN) which has been built by coupling the two types of our prior proposed models, namely Multiple Timescale Recurrent Neural Network (MTRNN) [3, 4] for behavior generation and Multiple Spatio-Temporal Neural Network (MSTNN) [5, 6] for recognition of dynamic visual image. It has been shown that MTRNN can learn to develop temporal hierarchy which enables compositional generation of action sequences while MSTNN, which is an extension of MTRNN, can develop adequate spatio-temporal hierarchy for recognition of complex dynamic visual patterns.

We expect that coupling of these two different functional networks can afford development of "synergetic" coordination among the aforementioned multiple cognitive processes. Especially, this objective can be achieved by allocating another dynamic network characterized by its slow timescale dynamics and long-ranged spatial connectivity on the top of those two networks for enabling them to interact deeply during iterative learning process.

This hypothesis is testified via simulation study of a humanoid robot, iCub for learning to perform two types of visually guided object manipulation tasks. In the first learning task, the iCub learns to grasp a visual object placed in different position and orientation while avoiding collision to a nearby obstacle. In the second learning task, the iCub observes human partner's gesture indicating one of two different objects located in front of the iCub. Then, the iCub attempts to grasp the indicated object. Both of the learning tasks are performed through repeated tutoring by the experimenter.

The paper attempts to clarify how the synergy can be developed via dynamic interaction among the proposed neurodynamic model, robot body, and environment in the tutoring process by conducting spatio-temporal analysis of the neural activity observed in the model network.

## 2 Related Works

There have been research trials on applications of deep learning or modified CNNs in generation of visually-guided robot behaviors. Celikkanat and colleagues [7] presented so-called that Recurrent Slow Feature Analysis (RSFA) by adding recurrent con-nection in the higher level of the original model of SFA developed by Wiskott and Sejnowski [8]. It was shown that the vision system can exhibit some visual object performance analogous to human infants as like tracking trajectory of objects and keeping their identities while occluded. Barros and colleagues [9] applied their development of so-called the Multi Channel Convolutional Neural Network (MCCNN) to human-robot interaction scenario. In MCCNN, multiple successive pixel frames in video stream for human movement is used to form basic motion representation with applying a Sobel operator in both horizontal and vertical directions which result in multiple channels of 3D visual information. Each CNN for each channel performs cubic convolution on the motion representation in the earlier level at each channel. In the final layer, movement categorical is obtained by integrating all CNN channels.

Although these studies mentioned above are interesting in terms of pioneering possible application of deep learning of dynamic vision in robotics domain, the part of dynamic vision perception has not been integrated to the one for motor generation in forming synergy.

## 3      The Proposed Method

### 3.1    Model Architecture

The proposed VMDNN model was composed of MSTNN for dynamic visual perception and MTRNN for behavior generation and attention control as shown in Figure 1. The MTRNN has been proposed for development of temporal hierarchy via learning for behavior generation by utilizing multiple timescale constraint imposed on network dynamics. The MSTNN is an extension of MTRNN for the purpose of developing spatio-temporal hierarchy for dynamic vision by utilizing the multiple spatio-temporal scales constraint imposed on the network dynamics. The MSTNN can be considered conceptually as integration of CNN [1] with spatial hierarchy and MTRNN with temporal hierarchy. On the top of the two networks, a dynamic network with slow timescale property, referred to as the PFC subnetwork was allocated. The activation in MSTNN propagated to MTRNN through PFC. Next, we look at more details of each subnetwork.

*MTRNN:* It was shown that MTRNN can learn compositional action sequences by developing temporal hierarchy in a dynamic neural network model which is composed of multiple levels of CTRNNs consisting of leaky integrator neural units with different time constant at each level [3]. The MTRNN part in VMDNN consists of subnetworks of the MTRNN-slow ($M_S$) assigned with the largest time constant, the MTRNN-fast ($M_F$) with smaller one and the outputs ($M_O$) with the smallest one. $M_O$ generates sequential outputs for behavioral pattern and attention control as will be described in details later.

The dynamic activation of *i*th neural unit in a subnetwork in the *l*th level can be computed by using the synaptic inputs from the subnetwork in the *k*th level as follows.

$$u_{li}^{t+1} = \left(1 - \frac{1}{\tau_l}\right) u_{li}^t + \frac{1}{\tau_l}\left[\sum_{j \in N_k} w_{il,jk} y_{jk}^t + b_{li}\right] \quad (1)$$

$$y_{li}^t = f(u_{li}^t) \quad (2)$$

where t is the time step, $\tau_l$ is the time constant of the *l*th level subnetwork, $N_l$ is the number of neural units in the *l*th subnetwork, $w_{ij}$ is the weight and *f(x)* = *1.7159×tanh(0.6667× x)*.

*MSTNN:* In the VMDNN model, the MSTNN was used to process sequence of pixel images that the robot perceives while conducting various types of tasks. It was shown that MSTNN can recognize complex dynamic visual image patterns by introducing leaky integrator neural units with different time constant at each layer of CNN [5, 6]. In general, the lower layer activity is constrained by short distance connectivity with smaller local kernels and fast time constant in the leaky integrator neural units while the higher layer activity is constrained by long distance connectivity with larger kernels and slow time constant in the neural units. The dynamic activation of the *k*th feature neural unit located in (x, y) position in the *l*th level subnetwork can be computed by using synaptic convolutional inputs from the *l-1*th level subnetwork as:

$$u_{lk}^{txy} = \left(1 - \frac{1}{\tau_l}\right) u_{lk}^{(t-1)xy} + \frac{1}{\tau_l}\left[\sum_{n=1}^{N(l-1)} \left(k_{lkn} * v_{(l-1)n}^t\right)_{xy} + b_{lk}\right] \quad (3)$$

$$v_{lk}^{txy} = f(u_{lk}^{txy}) \quad (4)$$

where $\tau_l$ is the time constant of the *l*th level subnetwork, * is the convolution operator, $N_l$ is the number of feature maps in the *l*th level subnetwork, $k_{lkn}$ is the value of the kernel, $b_{lk}$ is the bias and *f(x)* = *1.7159×tanh(0.6667× x)*. In the current VMDNN model, MSTNN consists of three levels of subnetworks: vision input ($V_I$), MSTNN-fast ($V_F$) and MSTNN-slow ($V_S$).

### 3.2 Action Generation Mode

On the onset of action generation, the internal states of all neural units were set with neutral values. Then, the VMDNN receives sequence of pixel frame in the vision input subnet ($V_I$). Each neuron's activity was computed from the fast subnet in MSTNN to the output subnet in MTRNN successively at each time step. $M_o$ in MTRNN generated arm/hand movement patterns and the visual attention control signals. The arm/hand movement patterns were generated by means of prediction of proprioception states at next step. The predicted proprioceptive states were fed to the motor control of iCub as next step target joint angles in arm and hand. The visual attention control signal in the proposed model was composed with two functions, namely 1) moving the head in 2 DoF such that visual objects or obstacles can be gazed at the center of the camera frame and 2) changing resolution of foveal vision. The softmax function was used for encoding the prediction outputs for the arm-hand proprioceptive states, the camera-head rotational angles, and the resolution of foveal vision.

### 3.3 Training

The training of the VMDNN was conducted by means of supervised training using the visuo-proprioceptive sequence pattern and the one for the level of visual focus obtained through repeated tutoring as the target teacher. First, the MSTNN part was pre-trained in order to enhance robust feature development in the lower visual level which is analogous to the evidence from developmental neuroscience [10, 11]. The pre-training of MSTNN was conducted by temporarily allocating the categorization output layer implemented by the softmax function on the top of PFC subnetwork. Then, the MSTNN was trained as a classifier in a delay response manner. A set of dynamic visual images was given to the MSTNN with a corresponding target labels, such as object orientation and position. By means of training the MSTNN to classify the given dynamic images into a different label, feature development in the lower visual levels of the VMDNN can be improved.

After the pre-training of the MSTNN, the temporary categorization layer was detached. Then the pre-trained MSTNN and MTRNN were trained together with their tight coupling through the PFC by allowing dense interactions between two processes of the forward computation for the prediction and the back-propagation of the prediction error generated. In the forward computation, the dynamic activation of neural units propagated from MSTNN, PFC to MTRNN for predicting next step outputs while visual inputs were fed into MSTNN as described previously. On the other hand in the BPTT (Backpropagation through time), the prediction error from MTRNN outputs was back-propagated to PFC through time. Therefore, it is expected that the functional structures in those three subnetworks would co-develop via dense interaction between them while utilizing the lower level visual feature structures developed in the pre-training period. Such co-developed functional structures would generate the higher level cognitive function of sequencing actions or switching attention as well as the lower level processing of precise but flexible visuo-motor coordination by organizing synergy between these different levels. Next, more details about the training methods are described.

The learning was conducted by following the BPTT [12] scheme. It is, however, noted that the way of computing the delta error in the VMDNN is different from the one in the conventional RNN because of the leaky integrator term in the forward activation dynamics defined in from Eq. (1) to Eq. (4). During the BPTT training, the connectivity weights in the whole networks except the pre-trained part in MSTNN were updated so that the prediction error $E$ represented by Kullback-Leibler divergence between the model output sequences ($y_i^t$) and the teaching sequences ($\bar{y}_i^t$) in Eq. (5) can be minimized.

$$\sum_t \sum_{i \in M_O} \bar{y}_i^t \log \frac{\bar{y}_i^t}{y_i^t} \tag{5}$$

The delta error at the $i$th unit $\partial E/\partial u_i^t$, either for an output unit or an internal unit, is recursively calculated from the following formula:

$$\frac{\partial E}{\partial u_i^t} = \begin{cases} \bar{y}_l^t - y_l^t + \left(1 - \frac{1}{\tau_i}\right)\frac{\partial E}{\partial u_i^{t+1}} & , \quad i \in M_O \\ \sum_{k \in N} \frac{\partial E}{\partial u_k^{t+1}} \left[\delta_{ik}\left(1 - \frac{1}{\tau_i}\right) + \frac{1}{\tau_k} w_{ki} f'(u_i^t)\right], otherwise \end{cases} \quad (6)$$

where $w_{ki}$ is a downstream connectivity weight from the $i$th unit to the $k$th unit, $\delta_{ik}$ is Kronecker delta and $f'$ is the derivative of the sigmoidal function. The detailed description of MTRNN and MSTNN including their learning scheme using the BPTT should refer to [3] and [5], respectively due to space limitation.

## 4 Experiments

We performed two types of tasks to verify the proposed model. In the first experiment (Experiment I), we examined the model's capability of coordinating visual recognition, attention switching, action preparation and generation by conducting a task in which a robot grasps an object placed in different positions and orientation while avoiding collision to a nearby obstacle. In the second experiment (Experiment II), we conducted the experiment for extending the robot task by additionally introducing human gesture recognition. In this task, the robot observes human partner's gesture that indicates one of two objects located in front of the robot. Then, the robot attempts to grasp the object indicated.

### 4.1 Robot Simulation Platform

In the current research, we conducted the experiments with using the iCub simulator [13]. Figure 2 illustrates the iCub simulator settings in both experiments. iCub is a small humanoid robot developed based on the embodied cognition hypothesis [14] and its simulator provides an environment in which physical interaction between iCub and predefined objects can be reconstructed with a good accuracy including visual inputs [13]. Several studies have shown the plausibility of using the simulator for robotics research [15-17].

*Simulation Robot Configuration:* To obtain the dynamic visual images, we used the iCub simulator's camera embedded in its left eye. Those images were first resized to 64 (W) × 48 (H), converted to grayscale and normalized to -1 to 1. In order to implement attention switching, we used two joints in the neck (pitch and yaw) to control iCub's head so that the robot can locate the obstacle or object at the center of the visual frame by moving its head. Furthermore, the robot was configured to give different resolution of visual frame to the network depending on the focus of foveal vision generated in the MTRNN outputs. In order to equip the robot with object manipulation skills, we used iCub's right arm that consists of 16 joints including 9 joints in the fingers. The proposed model directly controls the joint of the right arm (7 DoF) whereas the rest 9 joints of the fingers were controlled by the program that converts the network's outputs (level of grasping from open to close all fingers together in 1 DoF) to the actual joint values of robot's finger (9 DoF).

*Network Configuration:* Vision input subnet ($V_I$) contains 1 feature map which is the image input (64 × 48) obtained from the robot's left eye and the time constant of each neural unit was set to 1. MSTNN-fast subnetwork ($V_F$) contains 3 feature maps with the size of 22 × 18 and the time constant was set to 2. In MSTNN-slow subnetwork ($V_S$), there were 6 feature maps and the size of the feature maps was 8 × 7. The time constant was set to 30. In the PFC, 20 neurons were used and their time constants were set to 150 to exhibit slow timescale dynamics. For those three subnetworks ($V_F$, $V_S$, PFC) the size of kernels inputs were set to 22 × 14, 8 × 6, 8 × 7 respectively. The sampling factors which denote the amount of shift of the kernel in convolution operation were set to 2, 2 and 1 respectively. In $M_S$ as the slow subnet in the MTRNN, there were 30 neurons and their time constants were set to 10. $M_F$ consists of 50 neurons with the time constant set as 2. Finally, $M_O$ consists of 110 neurons with the time constant set as 1. Activation values in those neurons in $M_O$ were transformed to 11 analog values that were: robot's neck joints (2), right arm joints (7), level of grasping (1) and level of focus in foveal vision (1). Prior to the both experiments, the MSTNN were pre-trained to enhance robust feature development in the lower visual level as mentioned previously. In pre-training of the MSTNN, dynamic visual images received from robot's left eye while showing grasping actions were used as input data. Those actions are the ones that were acquired from human-guided experiences of grasping an object with different orientations at different positions when there was no obstacle. For the target label in pre-training, we used the position and orientation of the object so that the MSTNN was trained for classification of object's position and orientation with given dynamic visual images. Experiment I: Obstacle Avoidance and Object Manipulation in iCub Simulator

## 4.2 Experiment I: Obstacle Avoidance and Object Manipulation in iCub Simulator

*Task Design*: The task of the robot was to grasp and then to lift up the target object (red box in Figure 2) while avoiding the collision between the obstacle (yellow box in Figure 2). The overall flow of the task is as follows: First, the robot observes the task space on which the obstacle and the object are located. Then, the robot attends visually to the obstacle by moving its head so that the obstacle is located at the center of the visual frame. Once the obstacle is attended, the robot moves its arm by avoiding the obstacle. After the arm is located in the position that the arm does not collide with the obstacle, the robot attends the object by moving its head to locate the object at the center of the visual frame. Then, the robot moves its arm again to grasp the object. When the arm reaches the top of the object, the robot focuses on the center of the frame so the area which contains the object and robot's hand can be given to the network with a higher resolution (Please see Figure 3). The robot grasps the object and then lifts it up.

The robot was tutored to generate the aforementioned goal-directed actions via repeated learning of human-guided experience. The tutoring or training of the robot was performed by considering possible generalization in learning such as for position or orientation variations for visual objects or obstacles in some degree. During the training, the network learned 45 cases consisting of 3 obstacle locations (-0.22, -0.27 and -0.32 on X axis of the task space), 3 object locations (0.20, 0.25 and 0.30 on Y axis of the task space) and 5 object orientations (-45°, -22.5°, 0°, 22.5°, 45°). The object's location on Z axis of the task space was fixed throughout the experiment. The size of the obstacle was 0.01 (X) × 0.26 (Y) × 0.2 (Z) and the size of the object was 0.028 (X) × 0.1 (Y) × 0.07 (Z). During the testing, we examined the model's generalization capability by placing the obstacle and object randomly within the range of trained cases. For example, the obstacle was randomly placed the object between -0.22 to -0.32 on X axis. Meanwhile, the object was randomly located between 0.20 and 0.30 on Y axis. In addition, the object's orientation was randomly selected within the range of between -45° to 45°. We trained the network for 46,000 epochs with the pre-trained MSTNN subnetworks and the learning rate was set to 0.0001. The initial weights were randomly chosen between -0.025 to 0.025.

*Results:* After the VMDNN learned the training sequences, we examined the performance of the trained VMDNN for test behavior generation with 52 trials of unlearned situations (i.e. unlearned obstacle position, object position and orientation). The test performance was evaluated in terms of a success rate across the 52 trials. Each trial was evaluated as success when the robot successfully grasped and lifted the object and evaluated as failure otherwise. The result showed that the success rate for the unlearned case was 84.61%. 44 trials were successful among 52 total trials (Videos of the robot experiments are available at http://neurorobot.kaist.ac.kr/project.html).

Figure 4 depicts the activities of several neurons in the MSTNN-fast, PFC and MTRNN-fast subnetworks as well as the output sequences of two testing cases in which the orientation of the object was different (44° and -44°) while the other conditions were same. The numbers at the bottom of the figure indicates the sequences of subtask in the experiment 1 that are: (1) observing the task space, (2) attending to the obstacle, (3) moving the hand to the above of the obstacle, (4) attending to the target object, (5) reaching the hand to the above of the target object and (6) reaching the hand to the near the object with focusing it (7) grasping and (8) lifting. It was clearly shown that the MSTNN-fast and MTRNN-fast developed their neural activity faster whereas the PFC developed its neural activity slower. Especially, the neuronal activations in the PFC subnetwork were similar at the beginning ((1) ~ (3) in the figure) but they started to develop differently while the robot was attending to the object (4) in these two cases. In this stage, the object was located at the center of the view (i.e. fovea) and the level of focus remained as minimum. This result implies that the attention control mechanism which locates the object at the center was enough to conduct this task. In addition, MTRNN-fast showed different activity while the robot was moving its hand close to the objects located with different orientations. This result implies that the model network might prepare different ways of pre-shaping of the arm-hand posture depending on the perceived orientation of the object (44° and -44°).

Moreover, we also evaluated the model's generalization performance by varying the time constant of the PFC subnetwork (Figure 5). We trained the model for 46,000 epochs with setting the different time constant of the PFC level. For both training and testing, all the other conditions were set as the same and a total number of 52 trials were examined for each time constant condition. With the time constant of 150, we obtained the best generalization performance (84.61%) whereas the time constant of 30 showed the worst performance (61.53%). In general, generalization performance decreased with a smaller value set for the time constant. However, setting the time constant with a value more than 150 did not enhance the generalization performance furthermore. In sum, the result showed the importance of slower timescale dynamics of the PFC subnetwork in a particular range.

### 4.3   Experiment II. Human Gesture Recognition and Object Manipulation

*Task Design:* In this experiment, a standing object and a laying object were placed in the left side and in the right side with possible switching of their positions at each trial. The robot's task was to grasp one of these two objects that had been indicated by a human partner by gesture. The overall task flow was as follows: The robot first observes the human gestures that are displayed on the screen in the simulator (Figure 2.(b)). The human points one of two objects on the task space (pointing left or pointing right). Then, the robot attends to the task space where two objects are located. The robot, then, focuses its attention to the target object that has been pointed by the human gesture. Then, the robot grasps and lifts up the target object. It is noted that the way of grasping for these two objects are different. A standing object should be grasped from its side while a laying object should be grasped from the top. The dimension of a standing object was 0.028 (X) × 0.07 (Y) × 0.15 (Z) and a laying object was with 0.028 (X) × 0.15 (Y) × 0.07 (Z).

Similar to the experiment I, the robot was tutored to generate the aforementioned actions through repeated learning of tutored experiences in different situations. During the training, the robot learned 8 cases by combining 4 different configurations of two objects with two different actional situations of either the left side object or the right side object to grasp as indicated by the human. 4 different configurations of two objects were: the standing or laying object in the left side (-0.15 in *X*) and similarly, the standing or laying object in the right side (0.07 in *X*).

The video of human gesture consisted of 40 frames and they were displayed on the "screen" in the simulation environment. Video shootings of the same subject with 10 trials were randomly selected for each case of pointing left or right. Thus, each trial showed slightly different visual sequence of human gestures. The network was trained for 37,000 epochs with the pre-trained MSTNN and the learning rate was set to 0.0001. The initial weights were randomly chosen between -0.025 to 0.025.

It is noted that this experiment is still preliminary because it does not include tests for generalization for unlearned situations such as variations with position or orientation of the objects. Nevertheless, it is worthwhile reporting because the analysis of the neural activities in the whole network could clarify how the context dependent actions can be generated by coordinating correlated activities among local networks.

*Results:* The tests for regenerating the tutored actions showed that the robot was able to grasp the object which was pointed by the human partner on the screen. In addition, it was observed that the robot exhibited different ways of grasping depending on the types of the object (Videos of the robot experiments are available at http://neurorobot.kaist.ac.kr/project.html). Figure 6 illustrates the activities of several neurons in the MSTNN-fast, PFC and MTRNN-fast as well as the outputs of MTRNN when the robot was conducting two different actions: grasping the standing object in the left side and grasping the standing object in the right side by following the human's gesture. The other task conditions were the same between the two. The numbers at the bottom of the figure indicates the sequences of subtask in the experiment II that are: (1) observing human gestures, (2) attending to the task space, (3) observing the task space, (4) attending to the target object, (5) reaching the hand to the above of the object, (6) reaching the hand to the near the object with focusing it, (7) grasping and (8) lifting it.

Those neurons in MSTNN-fast and MTRNN-fast showed fast dynamics whereas neural activity in PFC was developed slowly as similar to Experiment I. Moreover, the neural activations of the PFC started with the same activation state between the two but were developed differently when the robot observed the human's gesture ((1) in Figure 6). This activity in the PFC continued to develop differently toward the end of the whole task.

In the meanwhile, the MTRNN-fast exhibited mostly the same activation profile until the attention was shifted to the task space (2). This implies that the observed gesture information was mainly kept in the PFC. In addition, when the arm approached near the object with focusing on it, the activity in the MSTNN-fast changed drastically followed by the change also in the MTRNN-fast. It is assumed that these changes in the peripheral levels enabled the hand to approach the objects adequately depending on the object types by focusing them.

In addition, we also examine how the different combinations of the human gestures and the object configurations affected the development of the neural activity in the PFC. Figure 7 shows the neural activation trajectories of 2 representative neurons in the PFC for four different combination cases that are: grasping the left standing object (red), grasping the left laying object (blue), grasping the right standing object (green) and grasping the right laying object (cyan). As can be seen from the figure, the trajectories of those 4 cases began at the same point, but they developed to nearby but two distinct states at the end of the human gesture (1) depending on either pointing left or right. Although while the robot was reaching its hand to the above of the object (5), all trajectories were relatively similar, the activation started to change significantly when the robot moved its hand toward the object with focusing it (6). This supports our assumption on usefulness of having attention control with focusing for generating object-directed action.

## 5    Discussion

The proposed model, VMDNN was built by coupling of the MSTNN for dynamic vision processing, the MTRNN for hierarchical behavior generation and attention control, and the PFC allocated in the top of these two subnetworks. It was shown through a set of simulated iCub experiments that the proposed model can cope with developments of contextual cognitive control of the attention switch and action generation by utilizing dynamic visual inputs temporally kept in the memory.

This aspect was clarified especially in the analysis of the experiment II that involves with the cognitive competency of delayed-response to stimulus. Our analysis depicted especially in Figure 7 and Figure 8 implies that the following dynamic mechanism has been developed in the model network through mastering of the adopted task. The dynamic visual image perceived for the demonstrated human gesture was abstracted in both spatial and temporal dimensions via hierarchical processing of the MSTNN. The PFC subnet allocated on the top of the MSTNN read the "intention" underlying the human gesture through the continuous inputs from the MSTNN and stored its content in the dynamic memory. The underlying mechanism can be accounted by gradual development of two distinct neural activities in the PFC depending on the perceived gestures of either indicating the left side object or the right one.

Then after, in the PFC, the "intention" of the human partner inferred from the visual stream was mapped into own actional intentions for execution of visual attention shifts, action sequence generation with precise visuo-motor coordination. The PFC performed this job by the top-down control of the MTRNN by feeding the output signal into it. It is interesting to note that the MTRNN activity was differentiated significantly only after the visual attention shifted from the human to the task space even though the PFC activity had been differentiated largely prior to this moment between the two distinct human's intention cases. When the intended object was focused with a higher resolution, precise visuo-motor coordination for grasping and lifting up the object was finally achieved by utilizing the information loop established through these three subnetworks.

Our crucial argument is that the aforementioned synergy for generating cognitive behavior was developed as the result of the proposed deep learning performed on the all subnetworks coupled together. The iterative computation consisting of the forward dynamics from the MSTNN, the PFC, to MTRNN and the error BPTT in the same pathway in the inverse direction during learning enabled the development of such coordinated dynamic structure in the whole network. Another essential argument is that such coordinated dynamic structure can be developed by utilizing spatio-temporal constraints in terms of timescale and spatial connectivity adequately imposed on the architecture of the model network. This argument corresponds to the resent neuroscience research on the system level connectome [18] which have shown that cognitive brain functions are developed by utilizing the anatomical constraints including connectivity among local regions and differentiation in timescales among those local regions.

## 6    Conclusion

The current study introduced a model named as the Visuo-Motor Deep Dynamic Neural Network (VMDNN) which can learn to generate cognitive behaviors of robots by coordinating multiple cognitive processes including visual recognition, attention switching, action preparation and generation. The simulation study on the model using the iCub simulator showed that synergetic coordination among the subnetworks can be developed when iterative learning is performed on the whole network built on coupling of those subnetworks. The future study will examine the scalability of the proposed model for various complex tasks by using physical iCub robots.

## Acknowledgement

This work was supported by Mid-career Researcher Program through NRF grant funded by the MSIP (2014R1A2A2A01005491).

## References

[1]     A. Krizhevsky, I. Sutskever, and G. E. Hinton, "Imagenet classification with deep convolutional neural networks," in *Advances in neural information processing systems*, 2012, pp. 1097-1105.

[2]     A. Hannun, C. Case, J. Casper, B. Catanzaro, G. Diamos, E. Elsen*, et al.*, "DeepSpeech: Scaling up end-to-end speech recognition," *arXiv preprint arXiv:1412.5567,* 2014.

[3]     Y. Yamashita and J. Tani, "Emergence of functional hierarchy in a multiple timescale neural network model: a humanoid robot experiment," *PLoS Computational Biology,* vol. 4, p. e1000220, 2008.

[4]     J. Tani, "Self-Organization and Compositionality in Cognitive Brains: A Neurorobotics Study," *Proceedings of the IEEE,* vol. 102, pp. 586-605, 2014.

[5]     M. Jung, J. Hwang, and J. Tani, "Multiple spatio-temporal scales neural network for contextual visual recognition of human actions," in *Development and Learning and Epigenetic Robotics (ICDL-Epirob), 2014 Joint IEEE International Conferences on*, 2014, pp. 235-241.

[6]     M. Jung, J. Hwang, and J. Tani, "Self-Organization of Spatio-Temporal Hierarchy via Learning of Dynamic Visual Image Patterns on Action Sequences," *PLoS ONE,* 2015.

[7]     H. Celikkanat, E. Sahin, and S. Kalkan, "Recurrent Slow Feature Analysis for Developing Object Permanence in Robots," presented at the IROS 2013 Workshop on Neuroscience and Robotics, Tokyo, Japan, 2013.

[8]     L. Wiskott and T. J. Sejnowski, "Slow feature analysis: unsupervised learning of invariances," *Neural Comput,* vol. 14, pp. 715-70, Apr 2002.

[9]     P. Barros, G. I. Parisi, D. Jirak, and S. Wermter, "Real-time gesture recognition using a humanoid robot with a deep neural architecture," in *Humanoid Robots (Humanoids), 2014 14th IEEE-RAS International Conference on*, 2014, pp. 646-651.

[10]    P. J. Hancock, R. J. Baddeley, and L. S. Smith, "The principal components of natural images," *Network: computation in neural systems,* vol. 3, pp. 61-70, 1992.

[11]   B. A. Olshausen and D. J. Field, "Emergence of simple-cell receptive field properties by learning a sparse code for natural images," *Nature,* vol. 381, pp. 607-609, 06/13/print 1996.

[12]   D. E. Rumelhart, J. L. McClelland, and P. R. Group, *Parallel distributed processing* vol. 1: MIT press, 1986.

[13]   V. Tikhanoff, A. Cangelosi, P. Fitzpatrick, G. Metta, L. Natale, and F. Nori, "An open-source simulator for cognitive robotics research: the prototype of the iCub humanoid robot simulator," presented at the Proceedings of the 8th Workshop on Performance Metrics for Intelligent Systems, Gaithersburg, Maryland, 2008.

[14]   N. G. Tsagarakis, G. Metta, G. Sandini, D. Vernon, R. Beira, F. Becchi*, et al.*, "iCub: the design and realization of an open humanoid platform for cognitive and neuroscience research," *Advanced Robotics,* vol. 21, pp. 1151-1175, 2007/01/01 2007.

[15]   M. Schlesinger, D. Amso, S. P. Johnson, N. Hantehzadeh, and L. Gupta, "Using the iCub simulator to study perceptual development: A case study," in *Development and Learning and Epigenetic Robotics (ICDL), 2012 IEEE International Conference on*, 2012, pp. 1-6.

[16]   A. G. Di Nuovo, D. Marocco, A. Cangelosi, V. M. De La Cruz, and S. Di Nuovo, "Mental practice and verbal instructions execution: A cognitive robotics study," in *Neural Networks (IJCNN), The 2012 International Joint Conference on*, 2012, pp. 1-6.

[17]   D. Marocco, A. Cangelosi, K. Fischer, and T. Belpaeme, "Grounding Action Words in the Sensorimotor Interaction with the World: Experiments with a Simulated iCub Humanoid Robot," *Frontiers in Neurorobotics,* vol. 4, p. 7, 05/31 2010.

[18]   O. Sporns, *Networks of the Brain*. Cambridge, MA: MIT press, 2011.

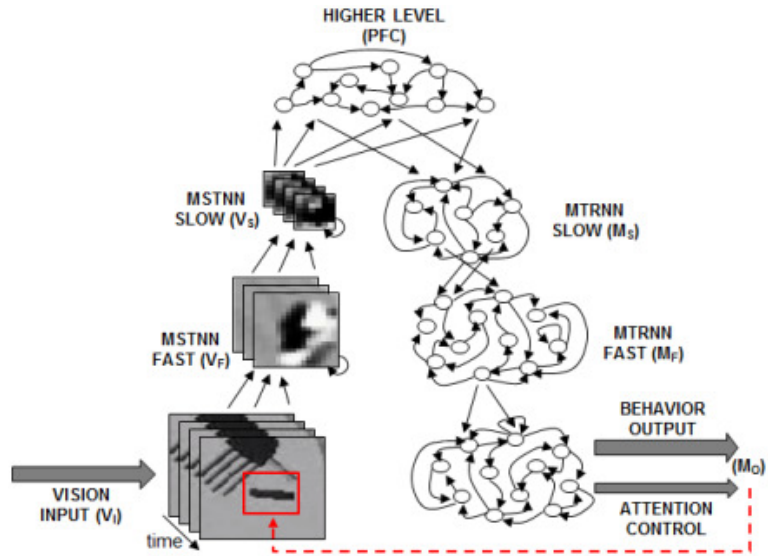

**Fig. 1.** The proposed VMDNN model. It consists of MSTNN [5, 6] for dynamic vision and MTRNN [3, 4] for behavior generation and attentional control, and the higher level network on the top of these two networks.

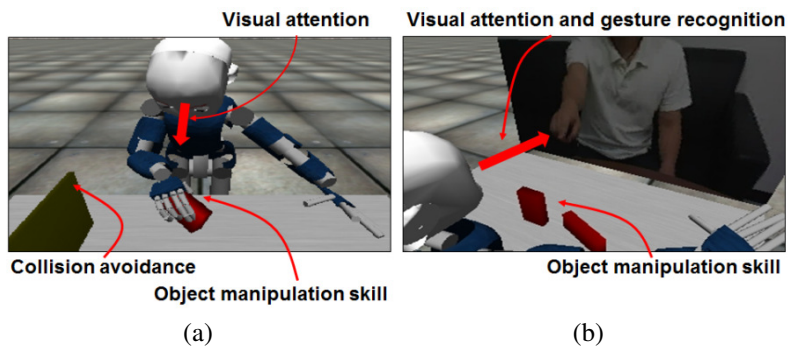

**Fig. 2.** iCub simulator settings. (a) Experiment I. The red box is the target object for grasping and the yellow box is the obstacle. (b) Experiment II. Human gesture video was displayed on the screen located in front of the robot and the robot grasps one of two objects located on the task space.

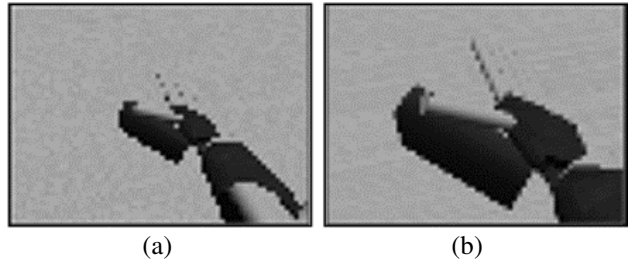

(a)                  (b)

**Fig. 3.** Examples of attention control in the proposed model. (a) Vision input when the model outputs the low level of focus in foveal vision and (b) vision input when the level of focus in foveal vision is high. Note that the visual scene containing the object and the hand at the center is given to the model with the higher resolution.

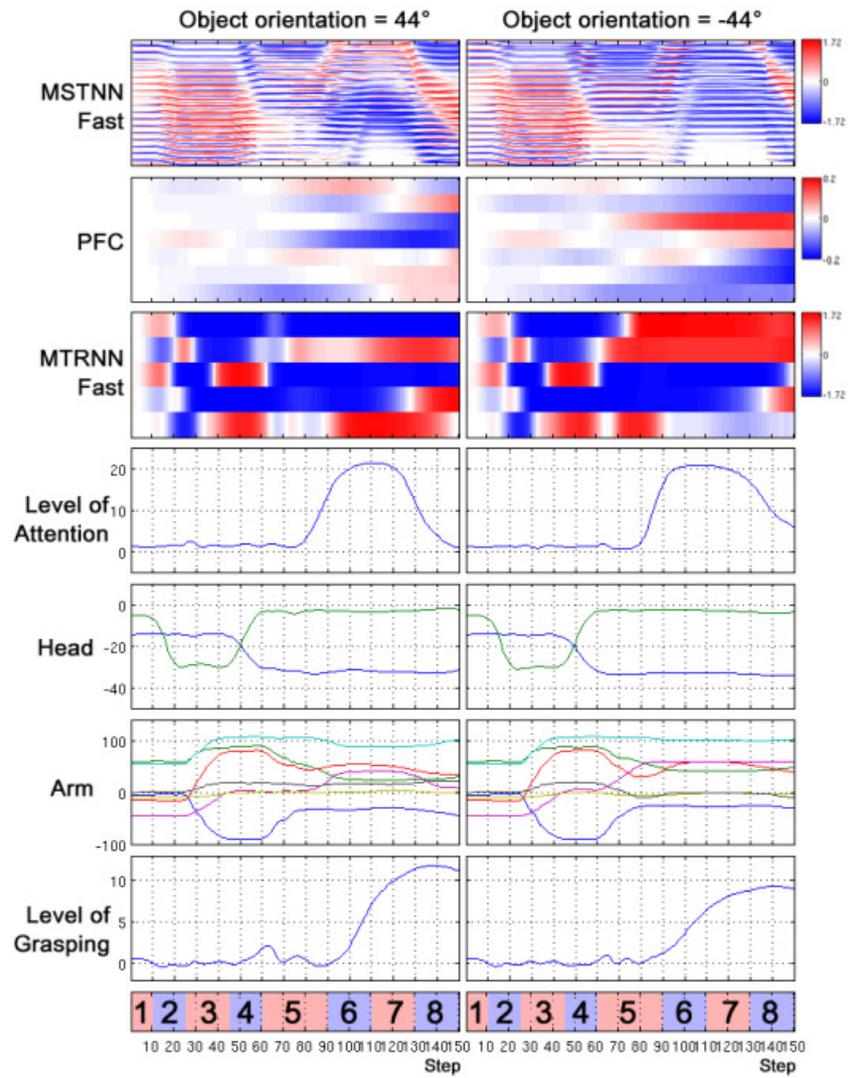

**Fig. 4.** An example of time developments of the neural activation of the three subnets and the model outputs in experiment I are shown. The X axis of each plot is time step and the Y axis of 3 upper rows indicate the individual neurons and the that of 4 bottom rows indicate the activation values.

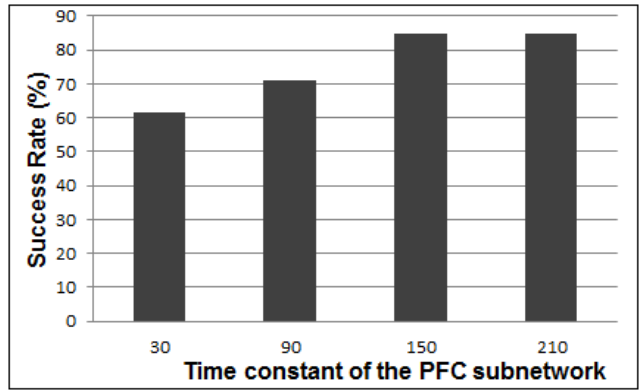

**Fig. 5.** Generalization performance of the 4 cases in which the time constant of the PFC subnetwork varied from 30 to 210.

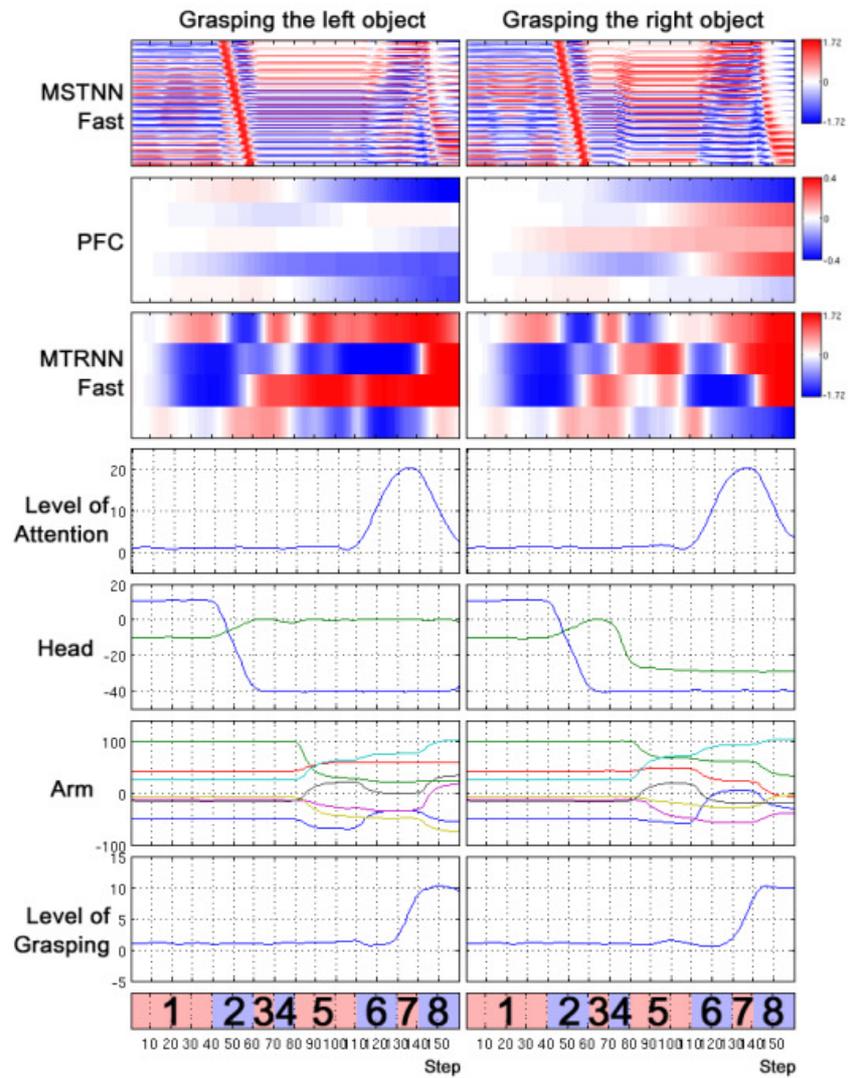

**Fig. 6.** An example of time developments of the neural activation of the three subnets and the model outputs in experiment II are shown. The X axis of each plot is time step and the Y axis of 3 upper rows indicate the individual neurons and the that of 4 bottom rows indicate the output values.

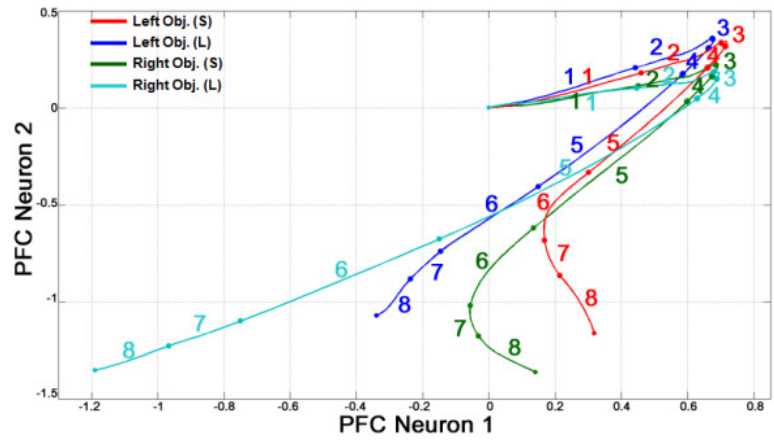

**Fig. 7.** The neuronal activation trajectories projected on 2-dimensional space for the four cases: grasping the left standing object (red), left laying object (blue), right standing object (green) and right laying object (cyan). The numbers next to the lines indicated the index of the current task that is equivalent to the ones described in the paper.